\documentclass[12pt,a4paper]{article}

\usepackage{arxiv}

\usepackage[utf8]{inputenc} 
\usepackage[T1]{fontenc}    
\usepackage{hyperref}       
\usepackage{url}            
\usepackage{booktabs}       
\usepackage{amsfonts}       
\usepackage{nicefrac}       
\usepackage{microtype}      
\usepackage{lipsum}
\usepackage{graphicx}
\usepackage{float}
\usepackage{amsmath}

\title{Fourier Transform Approach to \\Machine Learning I: Fourier Regression}

\author{
  Soheil Mehrabkhani \\
\\
  \texttt{soheil.mehrabkhani@alumni.uni-heidelberg.de} \\}

\begin{document}
\maketitle

\begin{abstract}
We propose a supervised learning algorithm for machine learning applications. Contrary to the model developing in the classical methods, which treat training, validation, and test as separate steps, in the presented approach, there is a unified training and evaluating procedure based on an iterative band filtering by the use of a fast Fourier transform. The presented approach does not apply the method of least squares, thus, basically typical ill-conditioned matrices do not occur at all. The optimal model results from the convergence of the performance metric, which automatically prevents the usual underfitting and overfitting problems. The algorithm capability is investigated for noisy data, and the obtained result demonstrates a reliable and powerful machine learning approach beyond the typical limits of the classical methods.
\end{abstract}


\vskip 0.7in

\section{Introduction}
Algorithm development to modeling a nonlinear relationship between response and the predictor is of great importance in supervised machine learning. Some of the interesting and powerful developed algorithms are polynomial regression [1-4], harmonic regression [5-7], smoothing methods[8-13] and k-nearest neighbors algorithm [1,14-17]. Many of the standard algorithms apply the method of least squares (LS) to determine the optimal model parameters. However, using the LS can cause issues related to the ill-conditioning of the required inverse matrix in the LS. Moreover, some of these methods require predefined hyperparameter to regularize the model and prevent underfitting and overfitting. Applying the smoothing functions with appropriate bandwidth are also widely spread, however, the choice of a predefined kernel function or information about the optimal bandwidth for arbitrary data are not always trivial.  Furthermore, some algorithms suffer from the too high computational effort. In principle, each of these algorithms has its own advantages and drawbacks and depending on the data and application, we have to decide, which algorithm to be applied. One challenging task is developing a fast fully automated general algorithm without requiring any hyperparameter or assumed additional information, which is not extracted from data.

In this work, we present an iterative algorithm based on the filtering the spectrum of the response function by the use of the discrete Fourier transform (DFT), which can be performed by a fast Fourier transform (FFT)[18-20].  The presented method does not apply the LS and thus, there is no need to compute the inverse of any matrix. To prevent the underfitting, the algorithm uses in each iteration the training data in spatial space to improve the performance of the last developed model. Subsequently, in frequency space and in each iteration, it filters the high frequencies to ensure the smoothness of the response function and to avoid the overfitting until the model is optimally trained. Because of the applied bandpass filtering, which is justified due to the data smoothness assumption, the smoothing kernel function is a $sinc$ function. The required bandwidth is not a fixed predefined or derived from cross-validation method, but it changes gradually until the used performance metric converges. Furthermore, the algorithm does not require any significant hyperparameter or additional information. Because in the algorithm the predictor and response variables are composed of the all training, validation, and test data together, the evaluation and prediction are updated simultaneously when the model is updated by the training data. In the next section, we dive into the detail of the theory behind the algorithm and how it actually works.   

\vskip 0.7in

\section{Learning by Iterative Fourier Transform }
\label{sec:headings}
\ Let us assume, there is a relationship between predictor values $x_i$ $(i\in \mathbb{N})$ and corresponding response values $y_i$, represented by a function $u$ as follows:
\begin{equation}
\ y_i=u(x_i).
\end{equation}
The standard approach in regression is finding a model represented by the function $u$ using a given predictor set (PS) $\{x_i\}$ and the corresponding known response set (RS) $\{y_i\}$, where the function u minimizes the residual sum of squares (RSS) between the predicted response values $\hat{y}_i=u(x_i)$ and true response values $y_i$. However, the actual goal is the generalization of the prediction model to unknown predictor sets. The common approach is randomly splitting the true training data $\{(x_i,y_i)\}$ into three subsets as training set $\{(x_i,y_i)\}_{tr}$ (TR), validation set $\{(x_i,y_i)\}_{va}$ (VA) and test set $\{(x_i,y_i)\}_{te}$ (TE). In principle, first the model must be trained, subsequently, validated and finally tested. Defining the set $\{x_i\}=\{x_i\}_{tr}\bigcup\{x_i\}_{va}\bigcup\{x_i\}_{te}$, all predictor values can be put together as a unified set to be applied
simultaneously in the algorithm and not separated. In other words, in our approach, all three steps will be accomplished
at once. 

One of the usual problems in the LS based regression methods is that the calculation of the response
 \ values $\hat{y}_i=u(x_i)$ requires a matrix inversion, which in some cases is ill-conditioned, thus, it cannot provide accurate results. In contrast to such methods, we apply a completely different approach using the DFT. Essentially, one common assumption in the standard DFT algorithms is uniformly sampled data [20]. However, in our algorithm, this is not a fundamental restriction because it may be easily removed by adding new predictor values to the set $\{x_i\}$, so that the new predictor set includes uniform distributed sample points. The sets $\{x_i\}$ and $\{\hat{y}_i\}$ must have an equal number of elements, therefore, we define $\hat{y}_i=0$ for all unknown response values corresponding to the validation, test and also newly added predictor points.

Due to applying inverse DFT (IDFT) in the algorithm, the function $u$ must be periodic [18]. To fulfill this assumption, the set $\{(x_i,\hat{y}_i)\}$ will be extended by adding its own mirror image.

The Fourier transform (FT) of the $\hat{y}$ results in its spectrum $\tilde{y}$:
\begin{equation}
\ \tilde{y}=\mathcal{F}\{\hat{y}\}.
\end{equation}
\ The bandpass filtering is performed by multiplication of the spectrum $\tilde{y}$ with a $rect$ function: 
\begin{equation}
\tilde{y}_f=\tilde{y} \ rect(\frac{f}{B}), 
\end{equation}
\ where $f$ and $B$ are frequency and the filtering bandwidth, respectively.
\ The inverse FT (IFT) of the filtered spectrum results in a modified function with the response values:
\begin{equation}
\ \hat{y}_f=\mathcal{F}^{-1}\{\tilde{y}\}=\mathcal{F}^{-1}\{\tilde{y}\ rect(\frac{f}{B}) \}=\mathcal{F}^{-1}\{\mathcal{F}\{\hat{y}\}\ rect(\frac{f}{B}) \}.
\end{equation}
\ According to the convolution theorem [21,22] and considering that the IFT of the $rect$ function is a $sinc$ function, Eq. (4) can be rewritten as follows:
\begin{equation}
\ \hat{y_f}=\mathcal{F}^{-1}\{\tilde{y}\} *\mathcal{F}^{-1}\{rect(\frac{f}{B}) \}=B\hat{y}* sinc(B\ x). \end{equation}

Depending on the bandwidth $B$, the convolution with the $sinc$ function makes the function $u$ smoother. The lower the bandwidth $B$, the smoother is the function $u$, however, the model will be too simple and underfit. On the other hand, for very high values of the $B$, the filtering has practically negligible influence on the smoothing and the model will be too complex and overfit. To find the optimal model, the iterative algorithm starts with the minimum value of the Bandwidth and gradually grows  until the optimal value is obtained. 
\ The filtered spectrum in the $n$-th iteration will be calculated based on Eq. (4) and by the use of a FFT algorithm: 

\begin{equation}
\ \hat{y_f}^{(n)}=\mathcal{F}^{-1}\{\mathcal{F}\{\hat{y}^{(n)}\}\ rect(\frac{f}{B^{(n)}}) \}.
\end{equation}

According to the DFT, the spacing step $\delta f$ in the frequency space  is equal to $1/L$, where $L$ is the width of the spatial window [18]. Because the convergence of the algorithm requires sufficient iterations, the actual steps must be smaller than from the DFT resulted theoretical minimum value $B_{min}=1/L$ . This is especially indispensable if the optimal bandwidth is comparable to the $B_{min}$. Without enough iterations per frequency steps, due to the lack of the iterations the algorithm would pass by the optimal point, and the convergence would occur, where the model is primarily affected by overfit. Furthermore, it may be expected, that for the optimal filtering bandwidth the performance of the algorithm converges. 

In principle, for a frequency step $\delta B=1/(mL)$ ($m \in \mathbb{N}, m>1$)  the filtering bandwidth $B_n$ does not change for the iteration numbers $n$ satisfying the condition $km \leq n<(k+1)m$ ($k\in \mathbb{N}$). For these iterations, the convergence condition imposes the invariance of the prediction performance for the optimal bandwidth. However, practically, some deviations can arise, thus the standard deviation $\sigma$ for the performance metric should be calculated for the iterations $n$ in the interval $km \leq n<(k+1)m$. We apply the $R^2$ score as performance metric, therefore, the standard deviation of the $R^2$ will be evaluated corresponding to the iteration sets with $m$ elements. If the standard deviation is smaller than a predefined minimum value $\sigma< \sigma_{min}$, the convergence is achieved and therefore the algorithm stops. These inequality is considered as the termination criteria of the algorithm.

According to the initial condition of the algorithm, the true values will be used for the training set and for all points outside the training set, the values are set to zero: 

\begin{equation}
\ \hat{y_i}^{(0)}=
  \left\{
    \begin{array}{l}
   \text{$y_i \quad  (x_i,y_i) \in TR$}\\
  \\
  \text{$0 \quad (x_i,y_i) \notin TR$}\\
    \end{array}
  \right.
\end{equation}

To train the model, the initial condition will be applied in the right side of Eq. (6) and after calculating the filtered values ${\hat{y}_{i,f}}^{(1)}$, the response values for the training set are replaced by the true values ${y_i}$. However, for all other points outside the training set, the  filtered values will be considered as the updated response values. By Repetition of the process, the response values in the $(n+1)$-th iteration $\hat{y}^{(n+1)}$ will be derived from the $n$-th iteration $\hat{y}^n$ by the use of Eq. (6) as follows:  

\begin{equation}
\ \hat{y_i}^{(n+1)}=
  \left\{
    \begin{array}{l}
   \text{$y_i \qquad  (x_i,y_i) \in TR$}\\
  \\
  \text{$\hat{y}_{i,f}^{(n)} \quad (x_i,y_i) \notin TR$}\\
    \end{array}
  \right.
\end{equation}

\vskip 1.0in

\section{Results}
\ To demonstrate the capability of the algorithm, we use  a data set function $u$ defined as follows:

\begin{equation}
\ y = u(x)= (\cos(0.1x^2) + \sin(8x)-\sin(1+0.1x^2)-\cos(1+8x))\exp(-0.01x^2)+ n,
\end{equation}

where the predictor variable $x$ is defined in the interval $[-25,25]$ and $y$ is the response variable. The data consists of the $N=512$ equidistantly distributed points. As described before, the uniform distribution is no fundamental distribution but just for the sake of simplicity, we deal in this work with
 an equidistant distribution of the data. The term $n$ is a Gaussian noise with the mean value $\mu_n = 0$ and the standard deviation $\sigma_n=0.1$.  
 
Figure (1) shows the true data according to Eq. (9) and the splitting the data randomly into training, validation and test sets. The training set includes $70\%$ and either the validation and the test sets contain $15\%$ of the total data.

\begin{figure}[H]
 \centering
 \includegraphics[width=16.5cm]{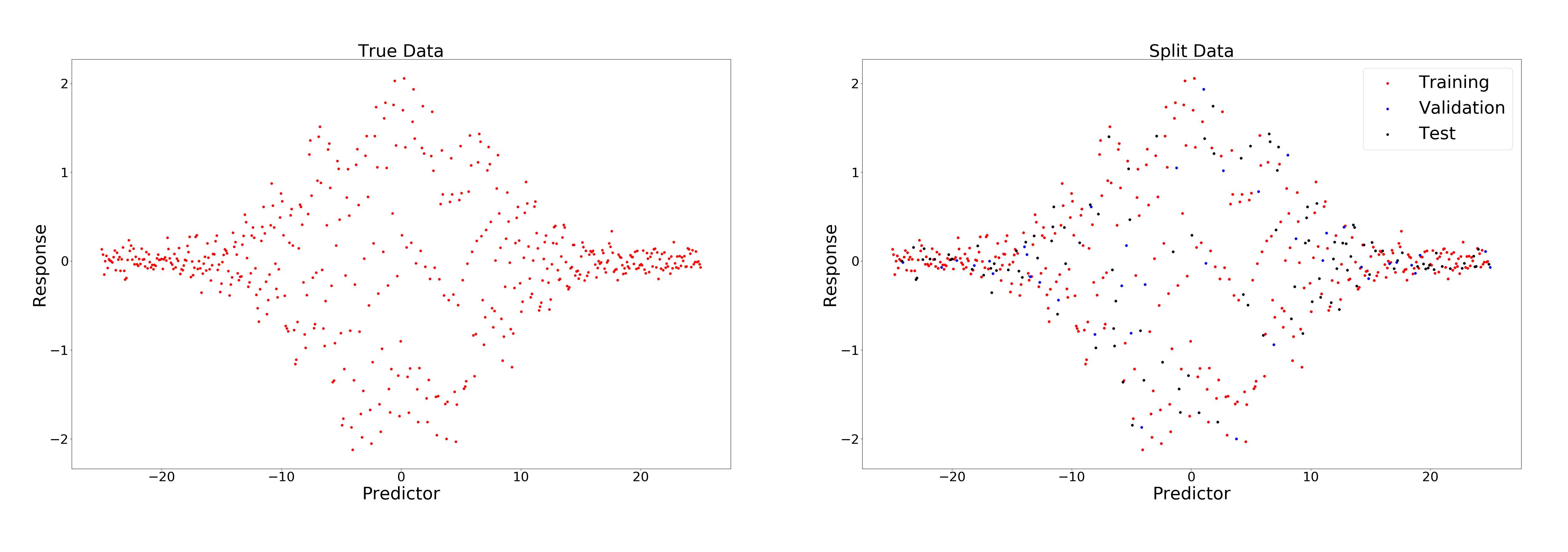}
 \caption{True and Split Data.}
 \label{fig:soheil1}
\end{figure}

As explained before, the original data set must be extended to become a periodic function by adding its own mirror image as shown in Fig. (2).

\begin{figure}[H]
 \centering
 \includegraphics[width=16cm]{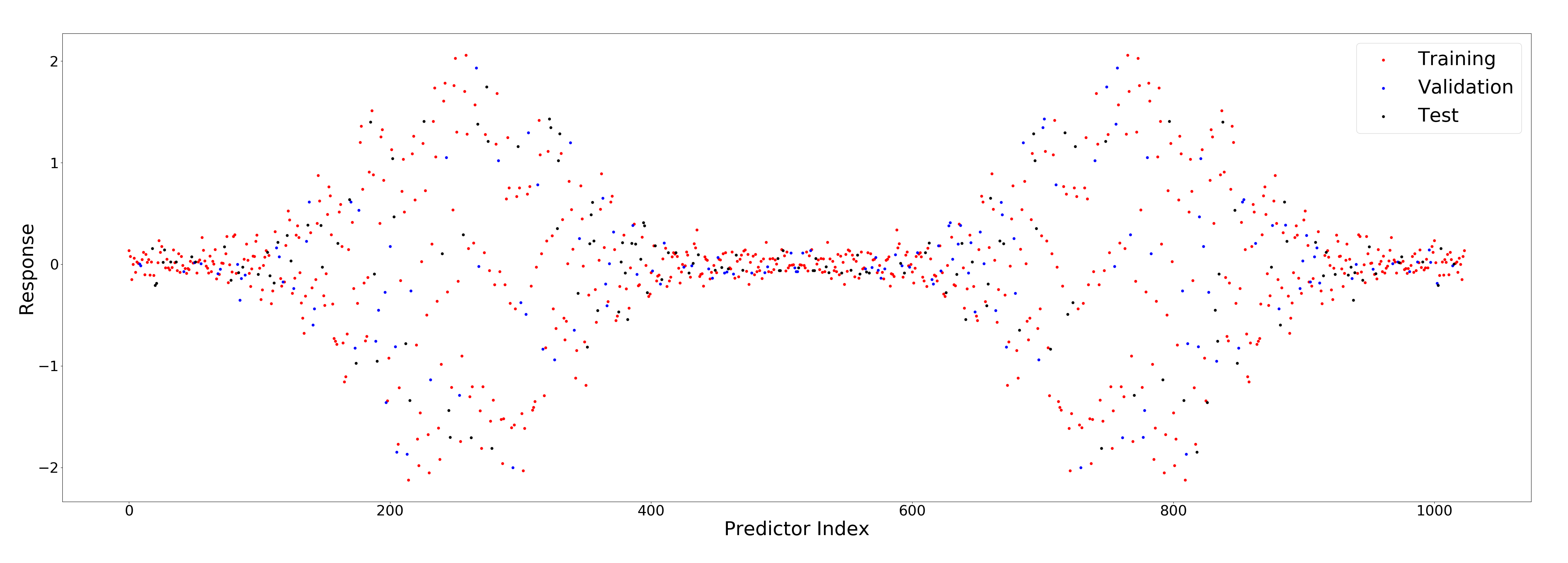}
 \caption{Extended Data. The extended data is symmetric and satisfies the required periodic condition}
 \label{fig:soheil2}
\end{figure} 

The training of the model and its evaluation start using the extended data set and by imposing the initial condition presented in Eq. (7) as illustrated in Fig. (3).  

\begin{figure}[H]
 \centering
 \includegraphics[width=16cm]{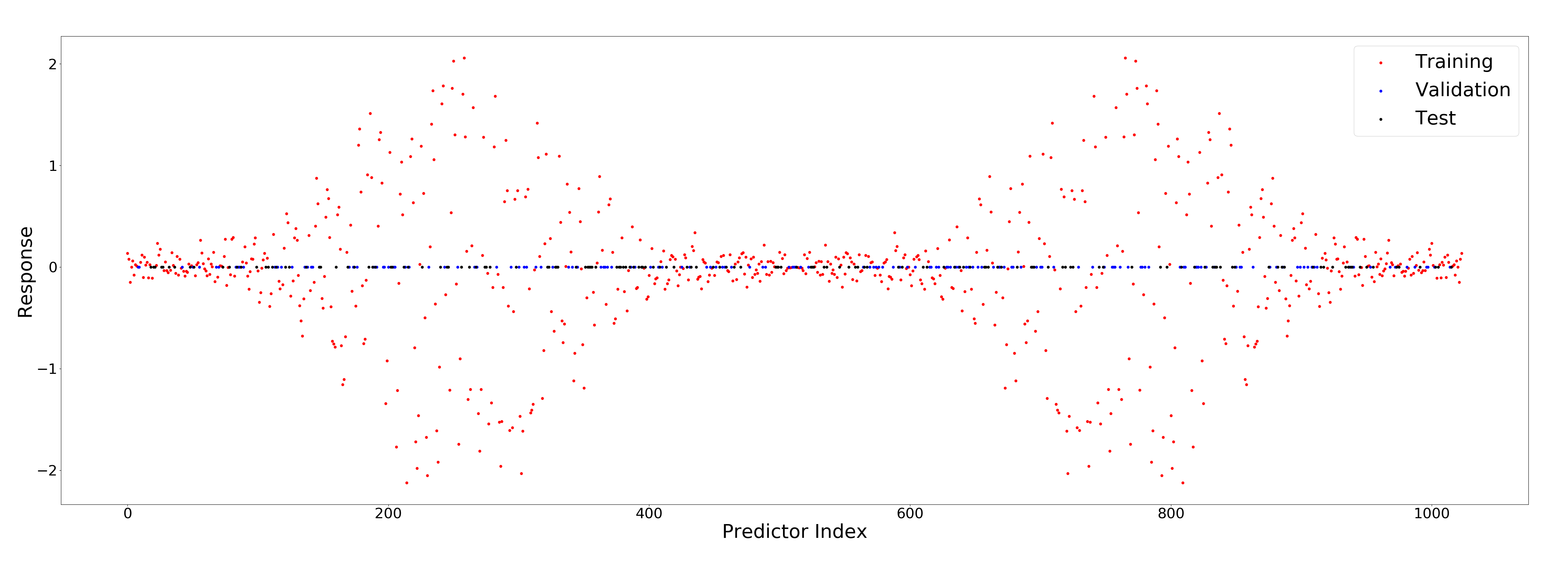}
 \caption{Initial Condition: All response values $\hat{y}$ outside the training set are set to zero}
 \label{fig:soheil3}
\end{figure} 

After applying the initial condition, the training of the model is performed based on Eq. (8). Figure (4) shows how the $R^2$ score changes with growing iteration number. The iterations will be repeated until the response values are sufficiently improved and the performance metric $R^2$ converges. As the convergence or termination criteria, the minimum standard deviation of the $R^2$ was set to $\sigma_{min}=0.0001$ with the iteration number $m=5$ per frequency step and total iteration number $N=100$

\begin{figure}[H]
 \centering
 \includegraphics[width=8cm]{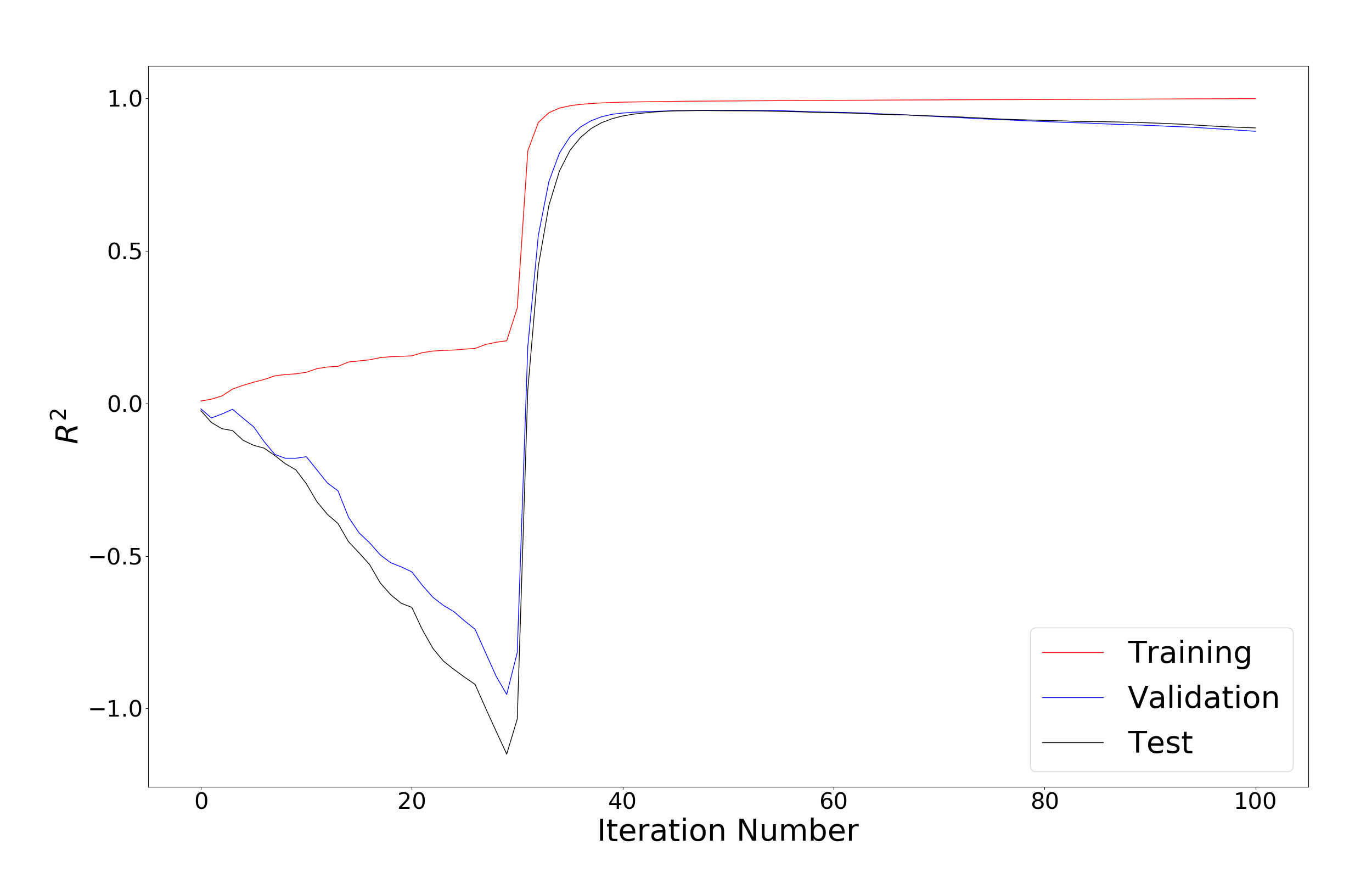}
 \caption{Dependency of the $R^2$ score from the iterations}
 \label{fig:soheil4}
\end{figure} 

As can be seen in Fig. (4), in the first iterations, the $R^2$ is too low and thus, the model performance is very poor. The main reason is that for these iterations, the filtering bandwidth is too low, and consequently, the trained model is too simple and underfit the data. In other words, the response values are primarily affected by the model itself and not the training data. Obviously, the model performance for the TR set increases slightly and for the VA and TE sets, it reduces rapidly.

These completely different behaviors refer to  Eqs (7) and (8). The calculated response values for the TR set are corrected in each iteration and thus, it can be expected that the model performance for the TR set should increase permanently, however, due to the underfitting, it improves very slowly. In contrast to the TR set, for the VA and TE sets, the algorithm starts with the initial values $\hat{y_i}^{(0)}=0$, which are very likely, not true but small.  Because the filtering bandwidth increases monotonically in the algorithm, the response values may grow but with probably wrong values, which can cause much lower model performance.

 After enough iterations, the bandwidth is sufficiently big, therefore it contains the most significant required frequencies for a good model describing the data. For the present example it happens, after about $30$ iterations. However, for better prediction result, the model needs more iterations until it reaches the convergence and the algorithm has to be stopped to prevent the overfitting problem. Due to the overfitting, the $R^2$ approaches to one for the training data, and it gradually reduces for the validation and the test data, which is a characteristic behavior if overfitting occurs.

\begin{figure}[H]
 \centering
 \includegraphics[width=16.5cm]{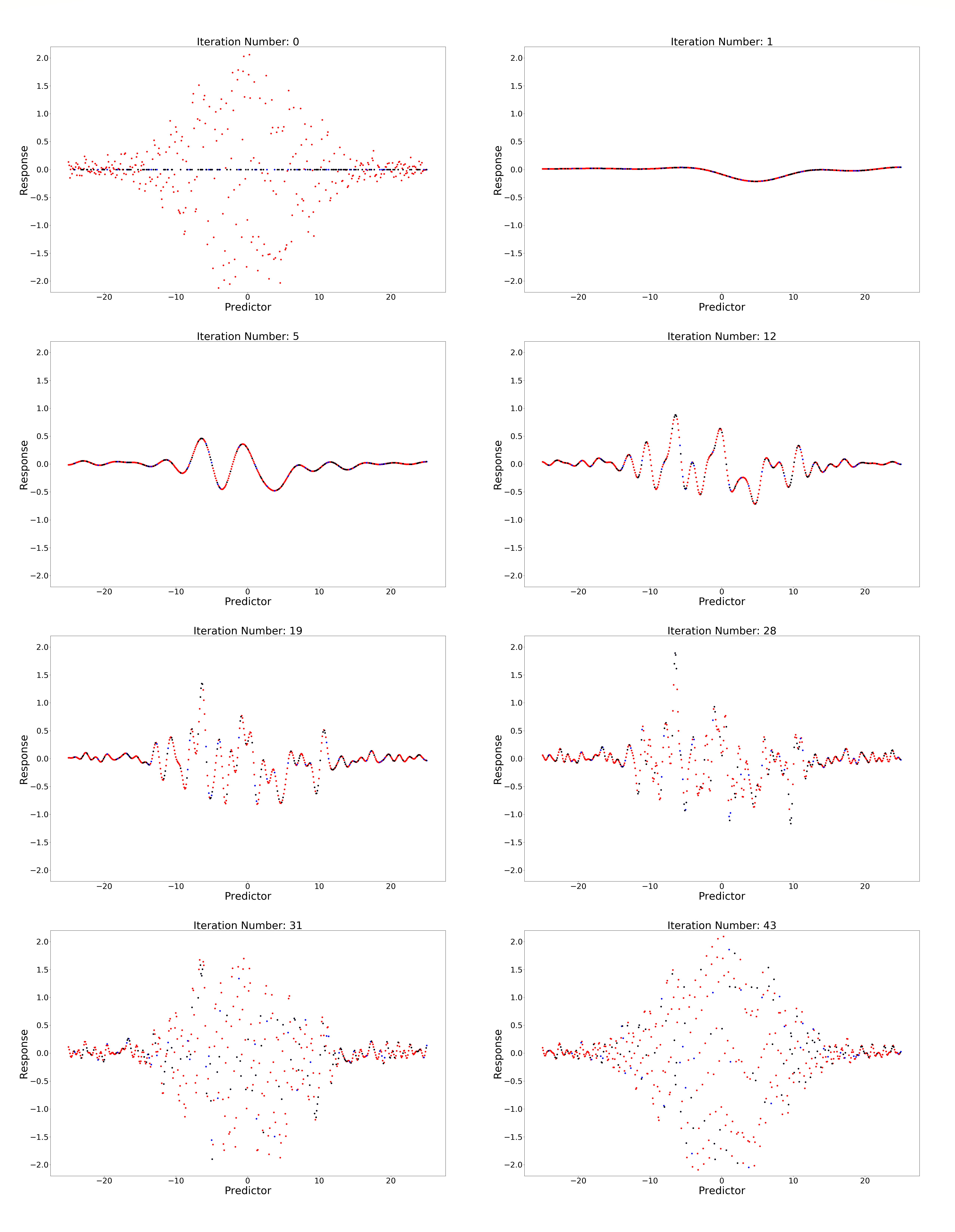}
 \caption{Training the model by the iterative algorithm}
 \label{fig:soheil5}
\end{figure} 

Figure (5) shows the model training based on the proposed algorithm, and how the model evolves dependent on the iterations. For the sake of visibility and because of the symmetric property only the first half of the data is shown.

\section{Conclusion}

We presented a supervising machine learning algorithm based on the DFT, which does not require the LS method, thus, it does not suffer from the typical restrictions of the LS based regression methods. Comparing to the smoothing approaches, the kernel function is a $sinc$ function, which is a direct result of the smoothing by bandpass filtering. The required bandwidth is not fixed but is determined automatically by the algorithm and dependent on the data itself.  Moreover,  this method does not need any hyperparameter or additional information for regularization and to prevent underfitting or overfitting problems. Because of the iterative nature of the algorithm, the enhancement of the model will be repeated until the convergence is achieved. One more advantage of the model is, that the training, evaluating and the test steps are combined together as one unified procedure. The capability of the method is demonstrated by the use of noisy non-periodic data. In the used example, the algorithm reaches $R^2=0.96$ after only $N=43$ iterations. The method can be applied for non-uniform sampled data and for multivariable data as well. We are convinced, that the developed iterative Fourier based learning algorithm can be extended to a very general learning approach for different machine learning areas like classification and clustering.

\bibliographystyle{unsrt}  


\end{document}